\documentclass[letterpaper, 10 pt, conference]{ieeeconf}  

\IEEEoverridecommandlockouts                              

\usepackage{graphics} 
\usepackage{epsfig} 
\usepackage{mathptmx} 
\usepackage{times} 
\usepackage{amsmath} 
\usepackage{amssymb}  
\usepackage{multirow}
\usepackage{graphicx}
\usepackage{float}
\usepackage{booktabs}
\usepackage{array}
\usepackage{xcolor}
\usepackage{colortbl}
\usepackage{pifont}

\definecolor{ForestGreen}{RGB}{34,139,34}
\usepackage{url}
\usepackage{bbding}
\usepackage{makecell}

\usepackage{threeparttable}

\usepackage{xspace}
\newcommand{\ie}{{\emph{i.e.}},\xspace}

\newcommand{\eg}{{\emph{e.g.}},\xspace}
\newcommand{\Eg}{{\emph{E.g.}},\xspace}
\newcommand{\etc}{etc.}

\newcounter{licomm}

\definecolor{Methodred}{RGB}{191, 3, 3} 
\definecolor{Secondblue}{RGB}{255, 127, 127}   

\newcommand{\goodnumber}[1]{{\color{Methodred}\textbf{#1}}}
\newcommand{\secondbestnumber}[1]{{\color{Secondblue}\textit{#1}}}
\newcommand{\method}{{\color{Methodred}TagaVLM}\xspace}

\title{\LARGE \bf

TagaVLM: Topology-Aware Global Action Reasoning for Vision-Language Navigation
}

\author{Jiaxing Liu$^{1,*}$, Zexi Zhang$^{1,2,*}$, Xiaoyan Li$^{1,\dagger}$, Boyue Wang$^{1}$, Yongli Hu$^{1}$, Baocai Yin$^{1}$
\thanks{The research project is partially supported by the National Natural Science Foundation of China (No.62506017, 62376014, 62441232).}
\thanks{*Equal contribution.}%
\thanks{$^{\dagger}$Corresponding author.}
\thanks{$^{1}$School of Information Science and Technology, Beijing University of Technology, China.}%
\thanks{$^{2}$Now at Imperial College London, U.K.}%
}

\begin{document}

\maketitle
\thispagestyle{empty}
\pagestyle{empty}

\begin{abstract}

Vision-Language Navigation (VLN) presents a unique challenge for Large Vision-Language Models (VLMs) due to their inherent architectural mismatch: VLMs are primarily pretrained on static, disembodied vision-language tasks, which fundamentally clash with the dynamic, embodied, and spatially-structured nature of navigation. Existing large-model-based methods often resort to converting rich visual and spatial information into text, forcing models to implicitly infer complex visual-topological relationships or limiting their global action capabilities. To bridge this gap, we propose \method({\color{Methodred}T}opology-{\color{Methodred}A}ware {\color{Methodred}G}lobal {\color{Methodred}A}ction reasoning), an end-to-end framework that explicitly injects topological structures into the VLM backbone. To introduce topological edge information, Spatial Topology Aware Residual Attention (STAR-Att) directly integrates it into the VLM's self-attention mechanism, enabling intrinsic spatial reasoning while preserving pretrained knowledge. To enhance topological node information, an Interleaved Navigation Prompt strengthens node-level visual-text alignment. Finally, with the embedded topological graph, the model is capable of global action reasoning, allowing for robust path correction. On the R2R benchmark, \method achieves state-of-the-art performance among large-model-based methods, with a Success Rate (SR) of 51.09\% and SPL of 47.18 in unseen environments, outperforming prior work by 3.39\% in SR and 9.08 in SPL. This demonstrates that, for embodied spatial reasoning, targeted enhancements on smaller open-source VLMs can be more effective than brute-force model scaling. The code can be found on our project page: \url{https://apex-bjut.github.io/Taga-VLM/}. 
\end{abstract}

\section{Introduction}

\begin{figure}[htbp]
\centerline{\includegraphics[scale=0.63]{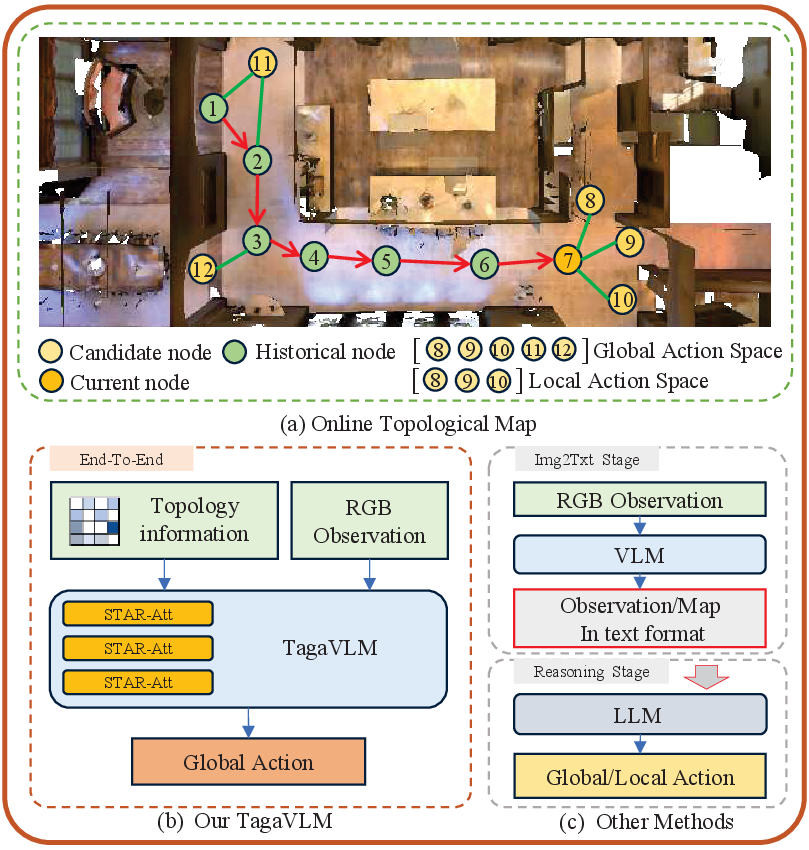}}
\caption{
Motivation of the proposed method.  Previous methods (c) usually employ a two-stage pipeline that uses VLMs to convert visual observations to text for decision-making in LLM, losing crucial visual information. In contrast, our \method (b) is an end-to-end paradigm that preserves VLM pretraining knowledge while incorporating online topological map information, enabling global action decisions (a) with backtracking ability.
}

\label{fig1}
\end{figure}
Vision-Language Navigation (VLN)~\cite{anderson2018vision} is a critical task for service robotics, requiring an agent to navigate to a target location in an unseen environment by following natural language instructions. This necessitates a deep understanding of language, visual perception, and their spatial grounding, enabling robust decision-making. 
Specifically, for VLN in discrete environments, the 3D environment is abstracted into a predefined connectivity graph. In this graph, each node represents a fixed, navigable viewpoint where the agent can observe its surroundings, while each edge connects adjacent nodes, defining a traversable path.

Due to the inherent characteristics of VLN in discrete environments, topological maps have played a crucial role in traditional VLN methods~\cite{ma2019regretful,wang2021structured,chen2022think,an2023bevbert}. SSM-VLN~\cite{wang2021structured} stores panoramic visual features of viewpoints in the nodes of topological maps. The edges between nodes encode orientation features corresponding to navigable viewpoints. DUET~\cite{chen2022think} stores both panoramic visual features and object features of viewpoints in the nodes of topological maps. The edges between nodes record the actual distances to navigable viewpoints. Topological-map-based methods are also successfully extended to continuous environments. \Eg ETPNav~\cite{an2024etpnav} employs a waypoint predictor to discretize continuous environments, thereby constructing online topological maps.
Compared with other types of methods, topological-map-based ones have achieved strong performance, mainly owing to two reasons. First, topological maps provide explicit visual-spatial correspondences, directly grounding visual features in the environment's spatial structure.
Second, it memorizes a global action space and enables the model to backtrack once an error occurs. Instead, other methods usually act on local space, which only consists of local navigable viewpoints directly connected to the current viewpoint.

In recent years, Large Language Models (LLMs) and Vision Language Models (VLMs) trained on massive data have demonstrated notable performance in various tasks, \eg natural language understanding~\cite{achiam2023gpt,team2024qwen2} and visual question answering~\cite{yang2023dawn,li2023blip-2}, even emerging certain spatial reasoning capabilities~\cite{yang2025thinking}. Consequently, many researchers have attempted to apply large models to VLN. However, as shown in Fig.~\ref{fig1}(c), most large-model-based methods, \eg NavGPT~\cite{zhou2024navgpt}, LangNav~\cite{pan2023langnav}, NavCot~\cite{lin2025navcot} \etc, 
employ a pretrained VLM to preprocess visual observations into text format, and use these textual descriptions for instruction tuning or zero-shot prompting of LLMs. 
However, the vision-to-text conversion and two-stage pipeline cannot sufficiently preserve and digest fine-grained visual information~\cite{chen2024mapgpt}. 
To overcome this limitation, some researchers directly utilize VLMs to construct end-to-end frameworks. \Eg NaviLLM~\cite{zheng2024towards} performs multimodal instruction tuning on large-scale multitask datasets, demonstrating strong multitasking capabilities. MapGPT~\cite{chen2024mapgpt} utilizes textural connectivity descriptions and adaptive planning mechanisms to help GPT-4V complete end-to-end navigation reasoning in a zero-shot manner. However, these methods ignore the gap between disembodied knowledge of pretrained VLMs and the embodied property of the VLN task, 
requiring the model to understand implicit visual-topological alignment passively and therefore increasing the learning difficulty. 
Additionally, with weak global topological understanding, these models are hard to backtrack decisions and are usually limited to local action decision space. 

To this end, a novel Topology-Aware Global Action reasoning framework, namely \method, is proposed. First, the proposed method designs a specialized image-text Interleaved Navigation Prompt (INP) to better align the textual and visual information from the same node in the topological graph and form input prompts with rich visual-textual context information. 
Second, the topological edge information is explicitly embedded in the network by Spatial Topology Aware Residual Attention (STAR-Att), which also keeps inheriting general knowledge in pretrained  VLM simultaneously. 
Finally, with the embedded topology map, all observed but unvisited viewpoints are included in the global action space at each navigation step, allowing the model to correct its navigation mistakes and further increase robustness effectively.
Our contribution can be summarized as follows:
\begin{itemize}

\item We introduce \method, an end-to-end VLN framework that architecturally embeds topological structures into the VLM backbone.

\item We propose two synergistic components: the INP that structures the input sequence to mirror the graph's node layout, and the STAR-Att mechanism that injects topological edge information directly into the self-attention layers, while preserving vital pretrained knowledge.

\item We provide strong evidence that, for embodied spatial reasoning, proper inductive bias is also critical besides sheer model scale. This principle is demonstrated as our \method-0.5B, with architecturally-injected topological priors, achieves competitive results, while the 7B version significantly outperforms previous, much larger proprietary models.

\end{itemize}

\begin{figure*}[htbp]
\centerline{\includegraphics[scale=0.45]{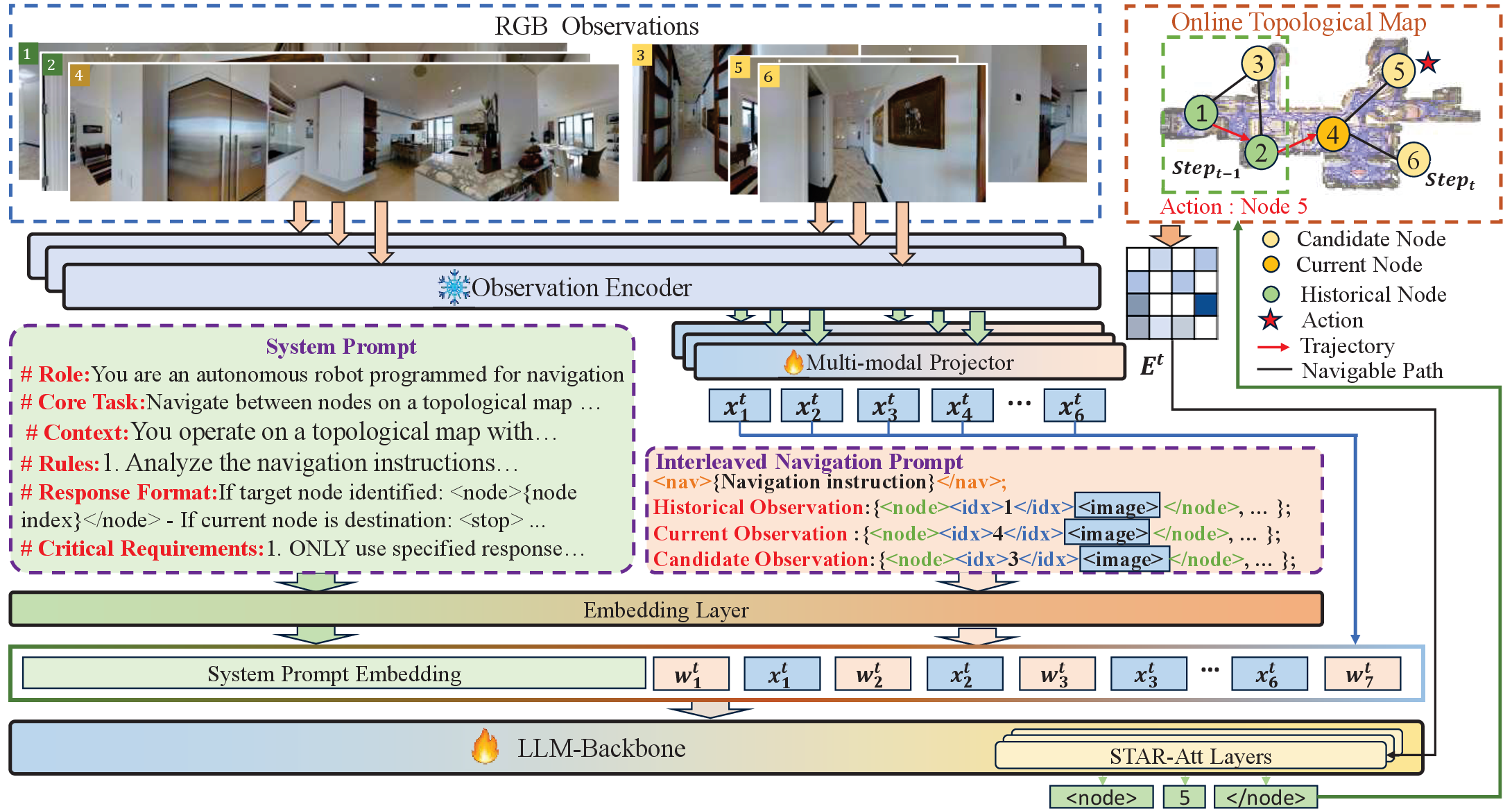}}
\caption{
Overview of the \method. The pretrained observation encoder and projector encode RGB observations from each node to the semantic space. Textual information containing navigation system prompts and navigation observation descriptions passes through embedding layers to obtain text embedding sequences. The observation feature sequences from each node are inserted into the text embedding sequences according to the corresponding \texttt{<image>} placeholder positions in the navigation observation descriptions, resulting in Interleaved Navigation Prompt Input. The LLM backbone is optimized with STAR-Att, enforcing the awareness of edge-level relationships by node pairwise distance matrices. Finally, the decisions are made in a global action space to guide the agent to move to the target node.
}
\label{fig2}
\end{figure*}
\section{Related Work} 



Navigating in unseen environments according to natural language instructions is an essential capability for autonomous robots. 
Previous VLN approaches mainly improve their methods from the following aspects: (1) \textbf{designing advanced network architectures}, \eg various Transformer-based architectures~\cite{qi2020object,hong2020language,an2021neighbor} and recently LLM and VLM based architectures; 
(2) \textbf{improving training paradigms}, \eg reinforcement and imitation learning~\cite{wang2019reinforced,tan2019learning}; 
(3) \textbf{augmenting data} 
for better generalization performance~\cite{wang2023scaling,hao2020towards,li2022envedit}. 
Among previous methods, multiple explicit memory mechanisms have played important roles, including episodic memory~\cite{qiao2022hop,chen2021history}, topology memory~\cite{chen2022think,an2024etpnav,zhao2022target} \etc\ These approaches have led the VLN models to exhibit strong performance. In addition, with the rapid development of LLM, large-model-based methods using instruction tuning~\cite{pan2023langnav,lin2025navcot,zheng2024towards} or zero-shot learning~\cite{zhou2024navgpt,chen2024mapgpt,long2024discuss} have also achieved promising results.

\textbf{Topology Maps for Vision-Language Navigation. } 
Metric maps constructed through SLAM~\cite{fuentes2015visual} have long dominated traditional visual navigation methods~\cite{huang2016visual}. Due to their ability to accurately record scene distributions, metric maps have also been explored in early Vision-Language Navigation approaches~\cite{anderson2019chasing}. However, heavy computational requirements~\cite{konolige2011navigation} limit their long-term planning capabilities~\cite{georgakis2022cross}. Consequently, researchers have proposed using graph-based topological maps as explicit memory storage during navigation, as these topological maps can efficiently align visual features with spatial information, enabling researchers to utilize them for environment pre-exploration~\cite{chen2021topological} and global navigation planning~\cite{ma2019regretful}. Taking advantage of the online topological maps, recent methods have achieved competitive performance on Vision-Language Navigation benchmarks, such as DUET~\cite{chen2022think} that employs dual-scale graph transformers or BEVBERT~\cite{an2023bevbert} that utilizes learnable metric maps.

\textbf{Large Models for Vision-Language Navigation. }
Recently, LLMs~\cite{achiam2023gpt,touvron2023llama,team2024qwen2} and VLMs~\cite{li2024llava,yang2023dawn,li2023blip-2} have emerged as research hotspots, 
given that they have learned extensive general knowledge. 
Most existing methods employ VLMs to convert visual observations into textual representations, which are then fed into LLMs for navigation reasoning, such as NavCot~\cite{lin2025navcot}, NavGPT~\cite{zhou2024navgpt}, LangNav~\cite{pan2023langnav}, and DiscussNav~\cite{long2024discuss}. Such two-stage systems suffer from significant visual information loss during the vision-to-text conversion phase~\cite{chen2024mapgpt}. Consequently, researchers have attempted to develop end-to-end large-model-based frameworks, for example, NaviLLM
~\cite{zheng2024towards} and MapGPT~\cite{chen2024mapgpt}. However, most of these methods wholly depend on the reasoning ability of LLM or VLM without additional global memory; therefore, they are usually constrained to local action spaces and are unable to perform backtracking actions. MapGPT~\cite{chen2024mapgpt} attempts to leverage textural connectivity
descriptions to enable GPT-4V to perform end-to-end navigation reasoning in global action spaces in a zero-shot manner. Nevertheless, research indicates that the large VLMs have bottlenecks in visual-spatial intelligence (\ie egocentric-allocentric transformation and relational reasoning)~\cite{yang2025thinking}. Therefore, relying solely on textual inputs without enhancing the model's intrinsic spatial perception forces the model to learn complex visual-topological relationships, increasing the learning difficulty.
Although existing large-model-based methods still exhibit various limitations, they are progressively narrowing the performance gap with traditional approaches while demonstrating strong generalization, zero-shot~\cite{chen2024mapgpt}, and multi-task~\cite{zheng2024towards} abilities that traditional methods lack. These advantages make them more aligned with real-world applications and encourage further exploration of large-model-based VLN research.

\section{Method}

To endow the VLM with inherent spatial understanding ability and topological-visual alignment power, while keeping its pretrained knowledge, in this work, a fully end-to-end Topology-Aware Global Action (\method) reasoning framework is proposed based on the pretrained large VLM, as shown in Fig.~\ref{fig2}. 
In the following subsections, we will elaborate on the four key components of our framework: (1) online topological map for global environment representation, (2) Interleaved Navigation Prompt for task adaptation, (3) Spatial Topology Aware Residual Attention for spatial knowledge embedded architecture, and (4) global action reasoning for correctable decision-making.


\subsection{Online Topological Map Environment Representation}
The online topological map, crucial in traditional VLN, can provide explicit visual-spatial correspondences at a low computational cost.
To supplement the spatial perception ability that the LLMs and VLMs are inherently lacking, an online topological map is employed, following the established configurations in traditional VLN methods~\cite{chen2022think}.

In the discrete environment setup for VLN, the entire navigation environment is configured as an unexplored undirected graph $\mathbf{G} = \{\mathbf{V}, \mathbf{E}\}$, where $\mathbf{V} = \{\mathbf{v}_i\}_{i=1}^K$ represents $K$ navigable nodes and $\mathbf{E}$ denotes the connecting edges between these nodes. Initially, the entire graph is invisible. As the navigation process progresses, the agent moves between navigable nodes along connected edges, and more nodes are gradually observed. At the navigation step $t$, the online topological map is denoted by $\mathbf{G}^t = \{\mathbf{V}^t, \mathbf{E}^t\} \subset \mathbf{G}$, where $\mathbf{V}^t=\{\mathbf{v}_i\}_{i=1}^{K^t}$ is the current node set with $K^t$ nodes, and $\mathbf{E}^t=\{d_{i,j}\}^{K^t\times K^t}$ is the current edge set, enumerating all possible connections between nodes. As shown in Fig. \ref{fig1}, three types of nodes are contained in $\mathbf{V}^t$: (1) historical nodes, (2) candidate nodes, and (3) the current node. 

\begin{figure}[htbp]
\centerline{\includegraphics[scale=0.7]{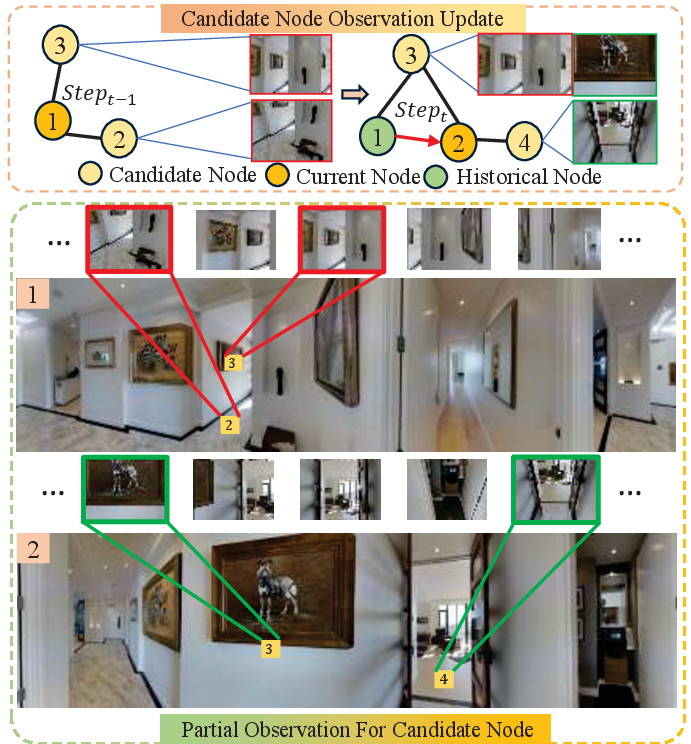}}
\caption{
In the navigation process, if an unvisited candidate node is observed multiple times at different positions, it is represented by stitching the images captured during all these observations. \Eg if $Node_3$ is observed both at position $Node_1$ and $Node_2$, the representation of $Node_3$ will be formed by concatenating the image of $Node_3$ observed at both $Node_1$ and $Node_2$.
}
\label{fig3}
\end{figure}

Each node $\mathbf{v}_i^t\in \mathbf{V}^t$ is represented by its observations. Each edge $d_{i,j}\in \mathbf{E}^t$ is expressed by the distance $d_{i,j}\in \mathbb{R}$ between the two nodes that it connects (\ie $\mathbf{v}_i^t$ and $\mathbf{v}_j^t$). Specifically, each of the historical nodes and the current node is represented by a panoramic image that is composed of images observed from 36 views at the current positions. Candidate nodes are unexplored and can only be partially observed from visited positions. Therefore, a candidate node is represented by a corresponding view where this node can be observed. As shown in Fig.~\ref{fig3}, when a candidate node is observed multiple times from different visited nodes, multiple observed views are concatenated to represent the candidate node.

At each step $t$, the current node $\mathbf{v}_c^t$ and its neighboring candidate nodes $\mathcal{N}(\mathbf{v}_c^t)$ are added to $\mathbf{V}^{t-1}$ to form the current node set $\mathbf{V}^t=\mathbf{V}^{t-1}\bigcup \mathcal{N}(\mathbf{v}_c^t)\bigcup \{\mathbf{v}_c^t\}$. $\mathbf{E}^{t}$ is updated accordingly, and the visual representations of current and candidate nodes are simultaneously updated based on new observations from $\mathbf{v}_c^t$.

\begin{figure*}[htbp]
\centerline{\includegraphics[scale=0.55]{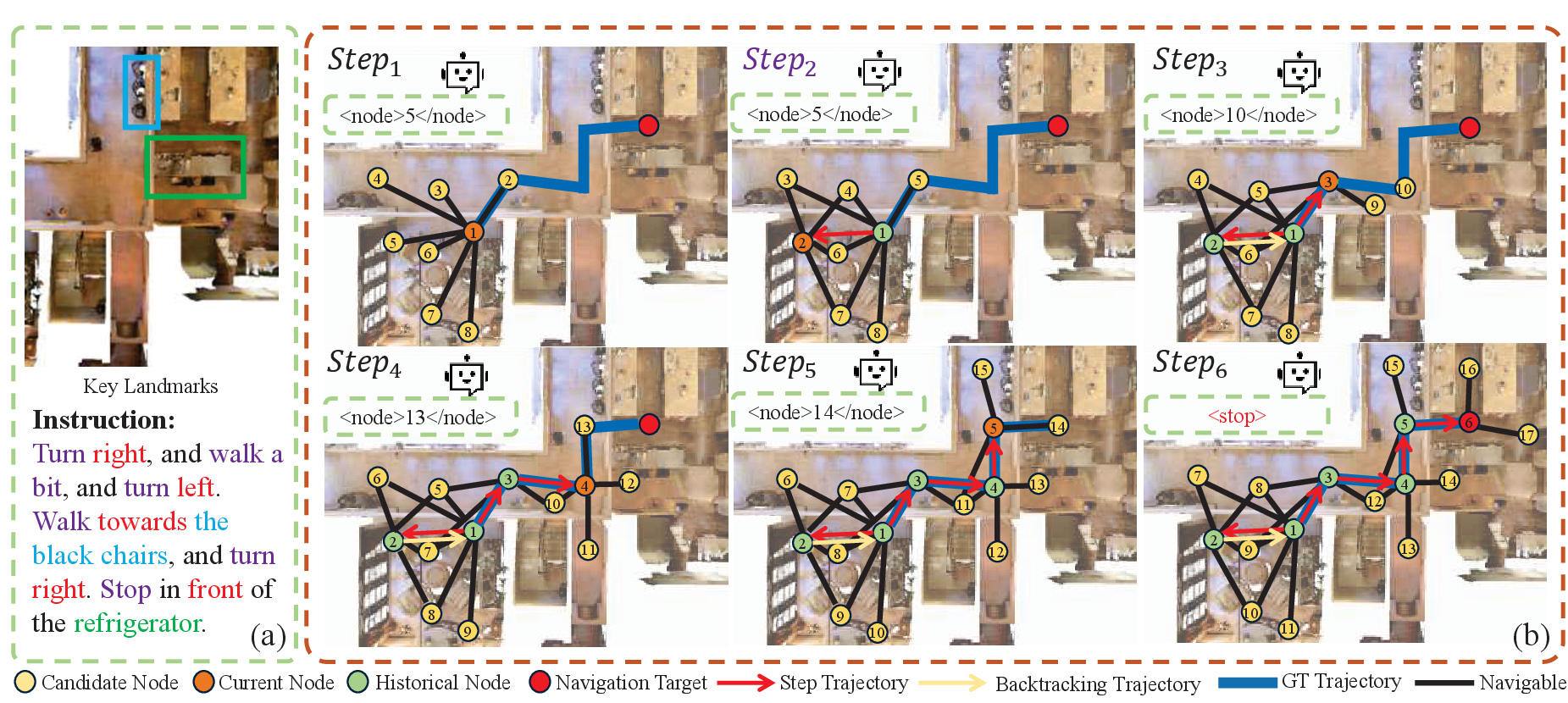}}
\caption{
A successful case demonstrates \method's spatial topological awareness and path correction ability. (a)  shows the navigation instruction containing two key landmarks: \textcolor{cyan}{black chairs} and \textcolor{ForestGreen}{refrigerator}. (b) Shows \method's 6 navigation steps from the starting node $Node_1$ to the target destination. In $Step_1$, due to the absence of landmark references, \method selected an incorrect direction. In $Step_2$, \method leveraged its spatial topological awareness capability and performed global action reasoning, selecting candidate node $Node_5$ of $Node_1$, backtracking from $Node_2$ to $Node_1$, and then moving to $Node_5$, successfully correcting the path. $Steps_{3-5}$ successfully followed the instruction by moving to the front of the \textcolor{cyan}{black chairs}, turning right, and moving to the front of the \textcolor{ForestGreen}{refrigerator}. Finally, in $Step_6$, \method made a stop decision and successfully reached the target destination.
}
\label{fig4}
\end{figure*}

\subsection{Interleaved Navigation Prompt}
 
Previous approaches \cite{chen2024mapgpt} include explicit alignment between visual observation IDs and node IDs in textual descriptions (\eg ``Place 1, corresponding to Image 1."), and sequentially arrange all visual observations at one end of textual prompts. 
This large contextual gap makes it difficult for the model to accurately associate a specific visual token with its corresponding text description. Consequently, the model's navigation reasoning capability is limited because it cannot properly align what it sees with the instructions and context it reads. 
In order to solve this problem, a text-image Interleaved Navigation Prompt (INP), which we denote as $\mathbf{P}^t_{INP}$, is designed to properly align visual observations with textual descriptions regarding navigation tasks, node sequences, and node types. Steps to form the proposed prompt are illustrated as follows.

First, the visual observations are encoded into visual tokens for further digesting in the LLM. Given $\mathbf{G}^t$ and the stored visual representations of each node $\mathbf{V}^t = \{\mathbf{v}_i^t\}_{i=1}^{K^t}$, an effective pretrained Vision Transformer (ViT)~\cite{dosovitskiy2020image} is employed as the observation encoder $\mathcal{E}(\cdot)$. After encoding, visual features $\mathbf{x}_i^t = \mathcal{E}(\mathbf{v}_i^t)$ are obtained for each node, where $\mathbf{x}_i^t$ represents a sequence of visual tokens with a length of 798. 
Then, a Multi-Layer Perceptron (MLP) is adopted as a multi-modal projector to map visual features into the input feature space of the LLM backbone, yielding $\tilde{\mathbf{x}}_i^t = \text{MLP}(\mathbf{x}_i^t)$.

Second, the visual features and textual features are assembled to form the $\mathbf{P}^t_{INP}$. As shown in Fig.~\ref{fig2}, the text in $\mathbf{P}^t_{INP}$ mainly includes the navigation instruction, node properties, and node IDs. They are first segmented into $K^t + 1$ parts according to the node IDs using observation placeholders \texttt{<image>} as delimiters. Then, each segment is encoded into textual features $\mathbf{w}_i^t$ ($i=1,\dots,K^t + 1$) through a pretrained embedding layer, where $\mathbf{w}_i^t$ comprises a sequence of multiple text tokens. Finally, the corresponding visual features $\tilde{\mathbf{x}}_i^t$ replace the placeholder, 
and $\mathbf{P}^t_{INP}$ can be formulated as, 
\begin{equation}
\mathbf{P}^t_{INP} = [\mathbf{w}_1^t, \tilde{\mathbf{x}}_1^t, \mathbf{w}_2^t, \tilde{\mathbf{x}}_2^t, \ldots, \mathbf{w}_{K^t}^t, \tilde{\mathbf{x}}_{K^t}^t, \mathbf{w}_{K^t+1}^t].
\end{equation}
This approach ensures that the visual features of each node contextually correspond to the node IDs and node types within the prompt, thereby strengthening the relevance between different representations of a single node. 

Finally, $\mathbf{P}^t_{INP}$ is concatenated with the system prompt $\mathbf{P}^t_{S}$ to form the complete input $\mathbf{P}^t$ of the LLM. The system prompt employs a structured organization, sequentially specifying the role, task, context, execution rules, response format, and constraint conditions for the LLM.


\subsection{Spatial Topology Aware Residual Attention (STAR-Att)}



While the Interleaved Navigation Prompt effectively aligns the visual and textual information for each individual node, it does not explicitly encode the spatial relationships between these nodes. To bridge this gap and enable the model to reason about the overall topological structure, we designed a Spatial Topology Aware Residual Attention (STAR-Att) mechanism. This mechanism embeds topological edge information directly into the network in a residual manner, achieving an explicit perception of visual-topological association information across all observed nodes. 

To design a topological-edge-related attention layer, first, the edge information is rearranged to form a token-wise affinity matrix. Assuming the current topological graph contains $K^t$ nodes, the edge set $\mathbf{E}^t$ can be represented by the current pairwise distance matrix $\mathbf{D}^t = \{d_{i,j}\}^{K^t \times K^t}$ as mentioned, where $d_{i,j} \in \mathbb{R}$ denotes the distance between the node $i$ and node $j$. When forming the prompt $\mathbf{P}^t$, 
we record the source of all tokens and generates a sequence of mask $\mathbf{M}^t = \{m_l^t\}_{l=1}^{L^t}$ ($m_l^t \in \{0,1\}$), which has the same length $L^t$ as the prompt $\mathbf{P}^t$. If the token originates from image features, $m_l^t$ is set to 1; otherwise, it is set to 0. Then, the original pairwise distance matrix $\mathbf{D}^t$ for nodes can be expanded  to an affinity matrix $\hat{\mathbf{D}}^t = \{\hat{d}_{p,q}\}^{L^t \times L^t}$ ($\hat{d}_{p,q}\in\mathbb{R}$) for tokens, 
\begin{equation}
\hat{d}_{p,q} = \begin{cases}
d_{i,j}, & \text{if } m_i^t = 1 \text{ and } m_j^t = 1 \text{ and } i \neq j \\
0, & \text{otherwise}
\end{cases},
\end{equation}
where $i$ and $j$ are the node indices, and when token $p$ and $q$ are both visual tokens, they are corresponding to node $i$ and $j$, respectively. 
Through the above steps, we obtain a token-wise affinity matrix where a higher value indicates a lower degree of correlation. 

Then, this matrix 
is fed into the proposed STAR-Att, together with the input prompt $\mathbf{P}^t$ to get the output features $\tilde{\mathbf{P}}^t$. The process of the STAR-Att can be formulated as, 
\begin{align}
\mathbf{S} &= \frac{\mathbf{P}^t\mathbf{W}_q(\mathbf{P}^t\mathbf{W}_k)^T}{\sqrt{d}} + \text{Linear}(-\hat{\mathbf{D}}^t),\\
\tilde{\mathbf{P}}^t &= \text{Softmax}\left(\mathbf{S}\right)\mathbf{P}^t\mathbf{W}_v,
\end{align} 
where 
$\mathbf{W}_q$, $\mathbf{W}_k$, and $\mathbf{W}_v$ denote the learnable parameters, and the Linear layer has a dimension of $1 \times N_h$, where $N_h$ corresponds to the number of attention heads in the original multi-head self-attention layers. The affinity matrix $\hat{\mathbf{D}}^t$ contributes to the bias term of the attention score $\mathbf{S}$. A longer distance between nodes is expected to result in a lower attention score in this mechanism, even if these nodes have similar visual features. 

We replace all multi-head self-attention layers with STAR-Att. Its core design—a learnable, per-head residual attention bias—implements the topological map as a flexible inductive prior, not a rigid constraint. This allows the model to dynamically weigh spatial cues against its powerful pretrained semantic knowledge, thus preserving its expressive capacity while gaining structured environmental awareness.

\begin{table*}[htbp]
\caption{Comparison with existing approaches on R2R dataset }
\vspace{-5pt}
\label{tab:results}
\begin{threeparttable}
\centering
\scalebox{0.95}{
\begin{tabular}{ll|ccccc|ccccc}
\toprule
\multirow{2}{*}{Method} & \multirow{2}{*}{Backbone} & \multicolumn{5}{c|}{Val Seen} & \multicolumn{5}{c}{Val Unseen} \\
\cmidrule{3-12}
& & TL & NE$\downarrow$ & OSR$\uparrow$ & SR$\uparrow$ & SPL$\uparrow$ & TL & NE$\downarrow$ & OSR$\uparrow$ & SR$\uparrow$ & SPL$\uparrow$ \\
\midrule
Seq2Seq~\cite{anderson2018vision} & LSTM &  11.33 & 6.01 & 53 & 39 & - & 8.39 & 7.81 & 28 & 21 & - \\
Speaker Follower~\cite{fried2018speaker} & LSTM &  - & 3.36 & 74 & 66 & - & - & 6.62 & 45 & 36 & - \\
HAMT~\cite{chen2021history} & Cross-Modal Transformer &  11.15 & 2.52 & - & 76 & 72 & 11.46 & 2.29 & - & 66 & 61 \\
DUET~\cite{chen2022think} & Cross-Modal Transformer &  12.32 & 2.28 & 86 & 79 & 73 & 13.94 & 3.31 & 81 & 72 & 60 \\
BEVBert~\cite{an2023bevbert} & Cross-Modal Transformer &  13.56 & 1.67 & 88 & 81 & 74 & 14.55 & 2.81 & 84 & 75 & 64 \\
ScaleVLN~\cite{wang2023scaling} & Cross-Modal Transformer &  13.24 & 2.12 & 87 & 81 & 75 & 14.09 & 2.09 & 88 & 81 & 70 \\
\midrule
NavGPT~\cite{zhou2024navgpt} & GPT-4\tnote{*} &  - & - & - & - & - & 11.45 & 6.46 & 42 & 34 & 29 \\
LangNav~\cite{pan2023langnav} & LLaMA2 (7B) & - & 7.4 & 40 & 32 & 28 & - & 7.1 & 45 & 34 & 29 \\
DiscussNav~\cite{long2024discuss} & GPT-4\tnote{*}  & - & - & - & - & - & 9.69 & \secondbestnumber{5.32} & 43 & 36.40 & 40 \\
NavCoT~\cite{lin2025navcot} & LLaMA2 (7B) & 10.08 & 6.46 & 48.38 & 41.33 & 38.43 & 9.95 & 6.26 & 48.11 & 40.23 & 36.64 \\
MapGPT~\cite{chen2024mapgpt} & GPT-4V\tnote{*} &  - & - & - & - & - & - & 5.62 & \secondbestnumber{57.9} & \secondbestnumber{47.7} & 38.1 \\
\midrule
\multirow{2}{*}{{\color{Methodred}TagaVLM} (ours)} & Qwen2 (0.5B) & 10.08 & \secondbestnumber{5.23} & \secondbestnumber{60.03} & \secondbestnumber{53.48}  & \secondbestnumber{50.4} & 9.8 & 5.57 & 55.09 & 45.72 & \secondbestnumber{41.91}\\
& Qwen2 (7B) &  10.16 & \goodnumber{4.71} & \goodnumber{64.15} & \goodnumber{55.53} & \goodnumber{53.05} & 9.7 & \goodnumber{4.97} & \goodnumber{60.2} & \goodnumber{51.09}  & \goodnumber{47.18}
\\
\bottomrule
\end{tabular}
}
\begin{tablenotes}
    \centering
    \footnotesize
    \item[*] Denotes proprietary models with undisclosed architecture and parameters, accessed via a black-box API.
\end{tablenotes}
\end{threeparttable}
\vspace{-5pt}
\end{table*}


\subsection{Global Action Reasoning}
For VLN in unseen environments, the ability to correct navigation paths during the navigation process is crucial. Therefore, we incorporate all observed but unvisited candidate nodes on the topological graph into the action space at each navigation step, termed the global action space. 

Specifically, our agent input includes the action space $\mathbf{A}^t = \{a^t_i\}_{i=0}^{N^t}$, allowing the agent to select the target viewpoint from $N^t+1$ options, where $i=0$ is defined as the stop action and $i>0$ corresponds to the current candidate nodes. At each time step, the agent selects $a^t_i$ and is required to output in a fixed format, such as: ``\texttt{<node>i</node>}'' or ``\texttt{<stop>}''. 
This global action space enables the model to perform global target selection. If the model predicts a non-adjacent historical or candidate node as its target, a shortest-path search algorithm~\cite{chen2022think} computes the low-level trajectory to that target, facilitating efficient path correction. For instance, as shown in Fig.~\ref{fig4}, with the global action reasoning ability, the proposed model is able to efficiently correct the decision error in the first navigation step. 

In the training process, the \method is finetuned with the single-step action prediction (SAP) task to align the instruction-trajectory paired data of VLN with the VQA training paradigm of VLMs. In each fine-tuning step, the model performs global action reasoning and predicts the target node index in a fixed format. 
The ground truth is directly obtained from the corresponding next node index in the ground truth trajectory provided by the R2R dataset. The training is conducted entirely in a teacher-forcing manner, where cross-entropy loss is computed between the predicted node index and the ground truth.

\section{Experiments}
We conducted experiments on \method to evaluate the performance on VLN compared with the previous state-of-the-art method and figure out the key components contributing to the performance gains. 

\subsection{Experiment Setup}

\textbf{Datasets.} Our experiments are conducted on the Matterport3D~\cite{anderson2018vision} simulator and evaluated on the R2R~\cite{anderson2018vision} dataset. 
To rearrange the dataset into SAP task for training, 14,093 trajectories from the R2R training split are segmented into discrete navigation steps. RGB visual observations, topological graph information, navigation instructions, and ground-truth actions are pre-extracted within the Matterport3D simulator, yielding a SAP training set with 80$K$ samples.

Furthermore, we leverage the data augmentation method \cite{wang2023scaling} to generate 500$K$ SAP samples (around 90$K$ trajectories) across 800 scenes from the HM3D \cite{ramakrishnan2021habitat} dataset. All the augmented data is used for training \method-0.5B. However, due to computational resource limitations, \method-7B is fine-tuned with only 200$K$ augmented samples. 

For testing, we utilize 1,021 navigation paths from the val seen split and 2,349 paths from the val unseen split in the R2R dataset. 

\textbf{Navigation Setup. }
In the  Matterport3D simulator, our agent is equipped with an RGB camera with a FOV of 60°, positioned at a height of 1.5m. Upon reaching each new node, the camera captures three sets of 12 views each, with pitch angles of 0°, +30°, and -30°, and yaw angles incremented by 30° steps. The panoramic view of each node is synthesized from these view sets. Candidate nodes are represented by a single view from the corresponding directional subset within the view collection.

\textbf{Implementation Details.}
We perform full fine-tuning on the parameters of the multimodal projector and the Qwen2~\cite{team2024qwen2} LLM backbone. The projector consists of two fully-connected layers with a hidden dimension of 1024, while the SigLIP~\cite{zhai2023sigmoid} visual encoder remains frozen. Our training employs a two-stage strategy: 1) pre-training for 12,500 steps on a mixture of R2R and augmented HM3D data, followed by 2) fine-tuning for 5000 steps on the R2R training data. For both stages, the Adam optimizer is used with a learning rate of 1e-5 and a batch size of 16.



\textbf{Evaluation Metrics. } The evaluation follows standardized metrics from the R2R dataset. In these metrics, Trajectory Length (TL) denotes average path length in meters; Navigation Error (NE) represents the average distance in meters between the agent’s final location and the target; Success Rate (SR) indicates the proportion of paths with NE less than 3-meter; Oracle Success Rate (OSR) is the SR given oracle stop policy; SPL is the SR penalized by Path Length.

\subsection{Comparison With Existing Approaches}

Previous methods can be broadly categorized into two types based on their model architectures: cross-modal-based and large-model-based methods. Cross-modal-based methods typically employ a smaller-scale LSTM or Transformer to either train from scratch or pretrain and then fine-tune for the VLN task. These methods achieve strong performance due to their task-specific design and dedicated training on navigation data. However, researchers continue to explore large-model-based approaches because of their inherent advantages in zero-shot generalization, multi-task capabilities, and superior language understanding, which make them more adaptable to diverse real-world scenarios. 
Among large-model-based methods, some retrain open-source LLMs specifically for navigation tasks, while others employ closed-source LLMs for training-free zero-shot inference. As shown in Tab.~\ref{tab:results}, our method surpasses the best-performing large model backbone methods across all five evaluation metrics in both splits. Notably, compared to MapGPT\cite{chen2024mapgpt}, our approach achieves an absolute improvement of 3.39\% in SR and 9.08 in SPL on the val unseen split. It is worth noting that, our 0.5B parameter model already outperforms most large-model-based methods and achieves comparable performance to state-of-the-art approaches with significantly larger parameter counts. It is important to note that our comparison to methods like MapGPT contrasts a fine-tuned specialist model (\method) with a zero-shot generalist model (GPT-4V). Our results demonstrate that by architecturally integrating the right inductive biases, smaller, open-source models can be fine-tuned to achieve superior performance and efficiency for a specific embodied task, presenting a compelling alternative to reliance on large, proprietary models.

In addition, it is an accepted fact that training on large-scale data can significantly improve model performance. However, owing to the limitation of computational resources, the amount of training data used for the proposed method is significantly smaller than that of NaviLLM\cite{zheng2024towards}, which uses over 1000$K$ trajectories and an additional 60$K$ QA samples. It prevents fair comparison between these methods, and therefore, NaviLLM is not compared in this work.
\begin{table}[H]
\caption{Ablation study on the R2R val unseen split.}
\vspace{-5pt}
\centering
\scalebox{0.87}{
\begin{tabular}{c|cccc|ccccc}
\toprule
&STAR-Att&INP&GA&AD&TL& NE$\downarrow$ & OSR$\uparrow$ & SR$\uparrow$ & SPL$\uparrow$ \\
\midrule
a &  \ding{56}&\ding{56}&\ding{56} &\ding{56} &13.85 & 9.05 & 27.37 & 17.28 & 13.01  \\
b & \ding{52}&\ding{56}&\ding{56} & \ding{56} &12.76&7.74&35.67 & 26.14 & 20.81      \\
c &\ding{52} & \ding{52} &\ding{56} &\ding{56} & 9.08 & 6.49 & 47.47 & 38.40 & 35.61  \\
d &\ding{52} & \ding{56} &\ding{52} &\ding{56} & 10.85 & 7.50 & 42.40 & 31.97 & 27.63  \\
e &\ding{52} & \ding{52} & \ding{52} &\ding{56} & 10.12 & 6.06 & 52.41 & 42.06 & 37.73  \\
f &\ding{52} & \ding{52} & \ding{52} & \ding{52} & 9.8 & \goodnumber{5.57} & \goodnumber{55.09} & \goodnumber{45.72} & \goodnumber{41.91}  \\
\bottomrule
\end{tabular}
}
\label{tab:ab}
\vspace{-5pt}
\end{table}
\subsection{Ablation Study}

To explore the effectiveness of key components in our approach and their impacts on navigation performance, we designed a series of ablation experiments to evaluate four critical components: Spatial-Topology Aware Attention (STAR-Att), Global Action (GA), Interleaved Navigation Prompt (INP), and  Augmentation Data (AD). All ablation experiments are conducted on the \method-0.5B model and the val unseen split of R2R dataset.

\textbf{STAR-Att vs. Text-Based Map. }
First, STAR-Att's impact is evaluated. In row (a) of Tab.~\ref{tab:ab}, all the essential components, including STAR-Att, are removed, and by solely finetuning the VLM to adapt the VLN task, it only achieves an SR of 17.28\%. When the standard multi-head self-attention is replaced by the proposed STAR-Att, row (b) showed a substantial performance improvement of 8.86\% in SR. 

\begin{table}[H]
\caption{STAR-Att vs. text-based map on the R2R val unseen split.}
\vspace{-5pt}
\centering
\scalebox{0.93}{
\begin{tabular}{c|cc|ccccc}
\toprule
&STAR-Att&Text-Based&TL& NE$\downarrow$ & OSR$\uparrow$ & SR$\uparrow$ & SPL$\uparrow$ \\
\midrule
a &\ding{56} &\ding{56} & 10.10 & 6.27 & 49.38 & 39.76 & 35.67 \\ 
b &\ding{56}& \ding{52} & 9.93 & 6.34 & 51.60 & 40.70 & 36.92  \\
c &\ding{52} &\ding{56} & 10.12 & \goodnumber{6.06} & \goodnumber{52.41} & \goodnumber{42.06} & \goodnumber{37.73}  \\
\bottomrule
\end{tabular}}
\label{tab:star}
\vspace{-5pt}
\end{table}

In addition, text-based topological map prompts are used as an alternative to STAR-Att for spatial structure perception. The results are shown in Tab.~\ref{tab:star} and all the experiments are carried out with both global action and Interleaved Prompt. Following MapGPT~\cite{chen2024mapgpt}, text-based topological map prompts are added, and as shown in row (b) of Tab.~\ref{tab:star}, the SR is increased by 0.94\% compared with the baseline in row (a). It illustrates the importance of topological information for the VLN. However, the text-based topological map achieves substantially lower performance improvements than the STAR-Att used in row (c), indicating significant challenges for understanding topological structures through textual representations.

\textbf{Interleaved Navigation Prompt.}
Compared with the baseline in row (b) of Tab.~\ref{tab:ab}, the model achieves large performance gains of 12.26\% in SR and 14.8 in SPL with the Interleaved Navigation Prompt, as shown in row (c). It highlights a critical insight: simply concatenating visual tokens creates a flat, unstructured input that is misaligned with the spatial nature of navigation. By interleaving prompts, we provide the essential scaffolding for the model to ground language in specific visual observations. More importantly, this structured sequence is what unlocks the full potential of STAR-Att, as it ensures that the spatial biases are applied to the correct and contextually aligned visual tokens, preventing the model from being limited in its navigation reasoning capability.

\textbf{Global Action Space vs. Local Action Space. } 
As shown in row (d) of Tab.~\ref{tab:ab}, the global action space is further added. A comparison between row (b) and (d) demonstrates that the global action space outperforms the local action space by a large margin of 5.83\% in SR and 6.82 in SPL. In comparison between row (c) and (e) with Interleaved Prompt, using the global action again shows apparent performance improvements. The global action space introduces backtracking capability that improves fault tolerance in navigation processes, enabling the agent to execute direct backtracking upon encountering navigation errors and select optimal directions for the next step, thus increasing the navigation SR and SPL.

\textbf{Augmentation Data. }
Finally, in row (f) of Tab.~\ref{tab:ab}, the impact of augmented data is assessed. By using the rich scene diversity provided by HM3D, an additional 500$K$ augmented data samples are utilized for the first-stage fine-tuning, which improves the model's generalization capability and leads to substantial performance gains in navigation tasks within unseen environments.

\section{Conclusions}

In this work, we introduce \method, a framework that directly tackles the mismatch between disembodied, pretrained VLMs and the embodied, spatial nature of navigation. Our approach architecturally embeds the topological graph into the VLM through two synergistic components: the Interleaved Navigation Prompt (INP), which enhances node-level visual-textual alignment, and our novel Spatial Topology Aware Residual Attention (STAR-Att), which explicitly integrates edge-level spatial relationships into the model's reasoning process. This holistic integration of topological structure empowers our model with global action reasoning ability and enables it to achieve state-of-the-art performance. Our results validate that enhancing VLMs with explicit spatial priors is a highly effective and efficient strategy for embodied reasoning, presenting a compelling alternative to relying solely on the implicit knowledge of larger-scale models. Future work will build on this strong baseline by scaling training on larger datasets, enriching STAR-Att with more complex geometric priors, and extending our framework to continuous control on physical robots.





\bibliographystyle{IEEEtran} 
\bibliography{IEEEabrv,ref}

@inproceedings{anderson2018vision,
  title={Vision-and-language navigation: Interpreting visually-grounded navigation instructions in real environments},
  author={Anderson, Peter and Wu, Qi and Teney, Damien and Bruce, Jake and Johnson, Mark and S{\"u}nderhauf, Niko and Reid, Ian and Gould, Stephen and Van Den Hengel, Anton},
  booktitle={Proceedings of the IEEE conference on computer vision and pattern recognition},
  pages={3674--3683},
  year={2018}
}

@article{fried2018speaker,
  title={Speaker-follower models for vision-and-language navigation},
  author={Fried, Daniel and Hu, Ronghang and Cirik, Volkan and Rohrbach, Anna and Andreas, Jacob and Morency, Louis-Philippe and Berg-Kirkpatrick, Taylor and Saenko, Kate and Klein, Dan and Darrell, Trevor},
  journal={Advances in neural information processing systems},
  volume={31},
  year={2018}
}

@inproceedings{hao2020towards,
  title={Towards learning a generic agent for vision-and-language navigation via pre-training},
  author={Hao, Weituo and Li, Chunyuan and Li, Xiujun and Carin, Lawrence and Gao, Jianfeng},
  booktitle={Proceedings of the IEEE/CVF conference on computer vision and pattern recognition},
  pages={13137--13146},
  year={2020}
}

@article{chen2021history,
  title={History aware multimodal transformer for vision-and-language navigation},
  author={Chen, Shizhe and Guhur, Pierre-Louis and Schmid, Cordelia and Laptev, Ivan},
  journal={Advances in neural information processing systems},
  volume={34},
  pages={5834--5847},
  year={2021}
}

@article{an2024etpnav,
  title={Etpnav: Evolving topological planning for vision-language navigation in continuous environments},
  author={An, Dong and Wang, Hanqing and Wang, Wenguan and Wang, Zun and Huang, Yan and He, Keji and Wang, Liang},
  journal={IEEE Transactions on Pattern Analysis and Machine Intelligence},
  year={2024},
  publisher={IEEE}
}

@inproceedings{chen2021topological,
  title={Topological planning with transformers for vision-and-language navigation},
  author={Chen, Kevin and Chen, Junshen K and Chuang, Jo and V{\'a}zquez, Marynel and Savarese, Silvio},
  booktitle={Proceedings of the IEEE/CVF Conference on Computer Vision and Pattern Recognition},
  pages={11276--11286},
  year={2021}
}

@inproceedings{chen2022think,
  title={Think global, act local: Dual-scale graph transformer for vision-and-language navigation},
  author={Chen, Shizhe and Guhur, Pierre-Louis and Tapaswi, Makarand and Schmid, Cordelia and Laptev, Ivan},
  booktitle={Proceedings of the IEEE/CVF Conference on Computer Vision and Pattern Recognition},
  pages={16537--16547},
  year={2022}
}

@article{anderson2019chasing,
  title={Chasing ghosts: Instruction following as bayesian state tracking},
  author={Anderson, Peter and Shrivastava, Ayush and Parikh, Devi and Batra, Dhruv and Lee, Stefan},
  journal={Advances in neural information processing systems},
  volume={32},
  year={2019}
}

@article{lin2025navcot,
  title={Navcot: Boosting llm-based vision-and-language navigation via learning disentangled reasoning},
  author={Lin, Bingqian and Nie, Yunshuang and Wei, Ziming and Chen, Jiaqi and Ma, Shikui and Han, Jianhua and Xu, Hang and Chang, Xiaojun and Liang, Xiaodan},
  journal={IEEE Transactions on Pattern Analysis and Machine Intelligence},
  year={2025},
  publisher={IEEE}
}

@inproceedings{li2023blip-2,
  title={Blip-2: Bootstrapping language-image pre-training with frozen image encoders and large language models},
  author={Li, Junnan and Li, Dongxu and Savarese, Silvio and Hoi, Steven},
  booktitle={International conference on machine learning},
  pages={19730--19742},
  year={2023},
  organization={PMLR}
}

@article{li2024llava,
  title={Llava-onevision: Easy visual task transfer},
  author={Li, Bo and Zhang, Yuanhan and Guo, Dong and Zhang, Renrui and Li, Feng and Zhang, Hao and Zhang, Kaichen and Zhang, Peiyuan and Li, Yanwei and Liu, Ziwei and others},
  journal={arXiv preprint arXiv:2408.03326},
  year={2024}
}

@inproceedings{yang2025thinking,
  title={Thinking in space: How multimodal large language models see, remember, and recall spaces},
  author={Yang, Jihan and Yang, Shusheng and Gupta, Anjali W and Han, Rilyn and Fei-Fei, Li and Xie, Saining},
  booktitle={Proceedings of the Computer Vision and Pattern Recognition Conference},
  pages={10632--10643},
  year={2025}
}

@inproceedings{zhou2024navgpt,
  title={Navgpt: Explicit reasoning in vision-and-language navigation with large language models},
  author={Zhou, Gengze and Hong, Yicong and Wu, Qi},
  booktitle={Proceedings of the AAAI Conference on Artificial Intelligence},
  volume={38},
  number={7},
  pages={7641--7649},
  year={2024}
}

@inproceedings{chen2024mapgpt,
	author = {Chen, Jiaqi and Lin, Bingqian and Xu, Ran and Chai, Zhenhua and Liang, Xiaodan and Wong, Kwan-Yee K.},
	booktitle = {Annual {Meeting} of the {Association} for {Computational} {Linguistics}},
	year = {2024},
	pages = {9796--9810},
	organization = {},
	title = {MapGPT: Map-{Guided} {Prompting} with {Adaptive} {Path} {Planning} for {Vision}-and-{Language} {Navigation}},
	volume = {},
}

@inproceedings{long2024discuss,
  title={Discuss before moving: Visual language navigation via multi-expert discussions},
  author={Long, Yuxing and Li, Xiaoqi and Cai, Wenzhe and Dong, Hao},
  booktitle={2024 IEEE International Conference on Robotics and Automation (ICRA)},
  pages={17380--17387},
  year={2024},
  organization={IEEE}
}

@article{pan2023langnav,
  title={Langnav: Language as a perceptual representation for navigation},
  author={Pan, Bowen and Panda, Rameswar and Jin, SouYoung and Feris, Rogerio and Oliva, Aude and Isola, Phillip and Kim, Yoon},
  journal={arXiv preprint arXiv:2310.07889},
  year={2023}
}

@inproceedings{zheng2024towards,
  title={Towards learning a generalist model for embodied navigation},
  author={Zheng, Duo and Huang, Shijia and Zhao, Lin and Zhong, Yiwu and Wang, Liwei},
  booktitle={Proceedings of the IEEE/CVF Conference on Computer Vision and Pattern Recognition},
  pages={13624--13634},
  year={2024}
}

@article{touvron2023llama,
  title={Llama: Open and efficient foundation language models},
  author={Touvron, Hugo and Lavril, Thibaut and Izacard, Gautier and Martinet, Xavier and Lachaux, Marie-Anne and Lacroix, Timoth{\'e}e and Rozi{\`e}re, Baptiste and Goyal, Naman and Hambro, Eric and Azhar, Faisal and others},
  journal={arXiv preprint arXiv:2302.13971},
  year={2023}
}

@article{an2023bevbert,
  title={BEVBert: Multimodal Map Pre-training for Language-guided Navigation},
  author={An, Dong and Qi, Yuankai and Li, Yangguang and Huang, Yan and Wang, Liang and Tan, Tieniu and Shao, Jing},
  journal={Proceedings of the IEEE/CVF International Conference on Computer Vision},
  year={2023}
}

@inproceedings{wang2023scaling,
  title={Scaling data generation in vision-and-language navigation},
  author={Wang, Zun and Li, Jialu and Hong, Yicong and Wang, Yi and Wu, Qi and Bansal, Mohit and Gould, Stephen and Tan, Hao and Qiao, Yu},
  booktitle={Proceedings of the IEEE/CVF international conference on computer vision},
  pages={12009--12020},
  year={2023}
}

@article{achiam2023gpt,
  title={Gpt-4 technical report},
  author={Achiam, Josh and Adler, Steven and Agarwal, Sandhini and Ahmad, Lama and Akkaya, Ilge and Aleman, Florencia Leoni and Almeida, Diogo and Altenschmidt, Janko and Altman, Sam and Anadkat, Shyamal and others},
  journal={arXiv preprint arXiv:2303.08774},
  year={2023}
}

@article{team2024qwen2,
  title={Qwen2 technical report},
  author={Team, Qwen},
  journal={arXiv preprint arXiv:2407.10671},
  year={2024}
}

@inproceedings{zhai2023sigmoid,
  title={Sigmoid loss for language image pre-training},
  author={Zhai, Xiaohua and Mustafa, Basil and Kolesnikov, Alexander and Beyer, Lucas},
  booktitle={Proceedings of the IEEE/CVF international conference on computer vision},
  pages={11975--11986},
  year={2023}
}

@article{yang2023dawn,
  title={The dawn of lmms: Preliminary explorations with gpt-4v (ision)},
  author={Yang, Zhengyuan and Li, Linjie and Lin, Kevin and Wang, Jianfeng and Lin, Chung-Ching and Liu, Zicheng and Wang, Lijuan},
  journal={arXiv preprint arXiv:2309.17421},
  year={2023}
}

@inproceedings{ma2019regretful,
	author = {Ma, Chih-Yao and Wu, Zuxuan and AlRegib, Ghassan and Xiong, Caiming and Kira, Zsolt},
	booktitle = {Computer {Vision} and {Pattern} {Recognition} ({CVPR})},
	year = {2019},
	pages = {6732--6740},
	organization = {},
	title = {The {Regretful} {Agent}: Heuristic-{Aided} {Navigation} {Through} {Progress} {Estimation}.},
	volume = {},
}

@inproceedings{wang2021structured,
	author = {Wang, Hanqing and Wang, Wenguan and Liang, Wei and Xiong, Caiming and Shen, Jianbing},
	booktitle = {Computer {Vision} and {Pattern} {Recognition} ({CVPR})},
	year = {2021},
	pages = {8455--8464},
	organization = {},
	title = {Structured {Scene} {Memory} for {Vision}-{Language} {Navigation}.},
	volume = {},
}

@inproceedings{qi2020object,
  title={Object-and-action aware model for visual language navigation},
  author={Qi, Yuankai and Pan, Zizheng and Zhang, Shengping and van den Hengel, Anton and Wu, Qi},
  booktitle={European conference on computer vision},
  pages={303--317},
  year={2020},
  organization={Springer}
}

@article{hong2020language,
  title={Language and visual entity relationship graph for agent navigation},
  author={Hong, Yicong and Rodriguez, Cristian and Qi, Yuankai and Wu, Qi and Gould, Stephen},
  journal={Advances in Neural Information Processing Systems},
  volume={33},
  pages={7685--7696},
  year={2020}
}

@inproceedings{an2021neighbor,
  title={Neighbor-view enhanced model for vision and language navigation},
  author={An, Dong and Qi, Yuankai and Huang, Yan and Wu, Qi and Wang, Liang and Tan, Tieniu},
  booktitle={Proceedings of the 29th ACM International Conference on Multimedia},
  pages={5101--5109},
  year={2021}
}

@inproceedings{wang2019reinforced,
  title={Reinforced cross-modal matching and self-supervised imitation learning for vision-language navigation},
  author={Wang, Xin and Huang, Qiuyuan and Celikyilmaz, Asli and Gao, Jianfeng and Shen, Dinghan and Wang, Yuan-Fang and Wang, William Yang and Zhang, Lei},
  booktitle={Proceedings of the IEEE/CVF conference on computer vision and pattern recognition},
  pages={6629--6638},
  year={2019}
}

@article{tan2019learning,
  title={Learning to navigate unseen environments: Back translation with environmental dropout},
  author={Tan, Hao and Yu, Licheng and Bansal, Mohit},
  journal={arXiv preprint arXiv:1904.04195},
  year={2019}
}

@inproceedings{li2022envedit,
	author = {Li, Jialu and Tan, Hao and Bansal, Mohit},
	booktitle = {Computer {Vision} and {Pattern} {Recognition} ({CVPR})},
	year = {2022},
	pages = {15386--15396},
	organization = {},
	title = {Envedit: Environment {Editing} for {Vision}-and-{Language} {Navigation}.},
	volume = {},
}

@inproceedings{qiao2022hop,
  title={Hop: History-and-order aware pre-training for vision-and-language navigation},
  author={Qiao, Yanyuan and Qi, Yuankai and Hong, Yicong and Yu, Zheng and Wang, Peng and Wu, Qi},
  booktitle={Proceedings of the IEEE/CVF Conference on Computer Vision and Pattern Recognition},
  pages={15418--15427},
  year={2022}
}

@inproceedings{zhao2022target,
  title={Target-driven structured transformer planner for vision-language navigation},
  author={Zhao, Yusheng and Chen, Jinyu and Gao, Chen and Wang, Wenguan and Yang, Lirong and Ren, Haibing and Xia, Huaxia and Liu, Si},
  booktitle={Proceedings of the 30th ACM international conference on multimedia},
  pages={4194--4203},
  year={2022}
}

@inproceedings{konolige2011navigation,
  title={Navigation in hybrid metric-topological maps},
  author={Konolige, Kurt and Marder-Eppstein, Eitan and Marthi, Bhaskara},
  booktitle={2011 IEEE International Conference on Robotics and Automation},
  pages={3041--3047},
  year={2011},
  organization={IEEE}
}

@inproceedings{georgakis2022cross,
  title={Cross-modal map learning for vision and language navigation},
  author={Georgakis, Georgios and Schmeckpeper, Karl and Wanchoo, Karan and Dan, Soham and Miltsakaki, Eleni and Roth, Dan and Daniilidis, Kostas},
  booktitle={Proceedings of the IEEE/CVF conference on computer vision and pattern recognition},
  pages={15460--15470},
  year={2022}
}

@article{fuentes2015visual,
  title={Visual simultaneous localization and mapping: a survey},
  author={Fuentes-Pacheco, Jorge and Ruiz-Ascencio, Jos{\'e} and Rend{\'o}n-Mancha, Juan Manuel},
  journal={Artificial intelligence review},
  volume={43},
  number={1},
  pages={55--81},
  year={2015},
  publisher={Springer}
}

@inproceedings{huang2016visual,
  title={Visual odometry and mapping for autonomous flight using an RGB-D camera},
  author={Huang, Albert S and Bachrach, Abraham and Henry, Peter and Krainin, Michael and Maturana, Daniel and Fox, Dieter and Roy, Nicholas},
  booktitle={Robotics Research: The 15th International Symposium ISRR},
  pages={235--252},
  year={2016},
  organization={Springer}
}

@inproceedings{dosovitskiy2020image,
	author = {Dosovitskiy, Alexey and Beyer, Lucas and Kolesnikov, Alexander and Weissenborn, Dirk and Zhai, Xiaohua and Unterthiner, Thomas and Dehghani, Mostafa and Minderer, Matthias and Heigold, G. and Gelly, S. and Uszkoreit, Jakob and Houlsby, N.},
	booktitle = {International {Conference} on {Learning} {Representations}},
	year = {2020},
	pages = {},
	organization = {},
	title = {An {Image} is {Worth} 16x16 {Words}: Transformers for {Image} {Recognition} at {Scale}},
	volume = {abs/2010.11929},
}

@article{ramakrishnan2021habitat,
  title={Habitat-matterport 3d dataset (hm3d): 1000 large-scale 3d environments for embodied ai},
  author={Ramakrishnan, Santhosh K and Gokaslan, Aaron and Wijmans, Erik and Maksymets, Oleksandr and Clegg, Alex and Turner, John and Undersander, Eric and Galuba, Wojciech and Westbury, Andrew and Chang, Angel X and others},
  journal={arXiv preprint arXiv:2109.08238},
  year={2021}
}
\end{document}